# White blood cell classification

Na Dong[1] · Meng-die Zhai[1] · Jian-fang Chang[1] · Chun-ho Wu[2]

1.   School of Electrical and Information Engineering, Tianjin University, 92 Weijin Road, Tianjin, China
2.   The Hang Seng University of Hong Kong, Hong Kong

**Abstract**

In recent years, conventional artificial method leads to low efficiency in the classification of blood cell, which requires professional completion. Therefore, the classification process is increasingly dependent on artificial intelligence. The traditional image classification method needs to extract a large number of features. Redundant features cause the recognition speed to be slow, and influence the recognition effect. To address these problems and obtain the highest recognition accuracy with the least number of features, this paper proposes a machine learning method based on feature selection algorithm for blood cell classification. Firstly, we introduced classification and regression trees (CART) for cell feature selection, which reduces the dimension of input feature attributes. Subsequently, particle swarm optimization (PSO) was used to optimise the hyper parameters of support vector machine (SVM) in this paper, making the SVM model better for classification. Finally, the Herlev dataset was introduced to verified the classification performance. The experimental results show that the proposed algorithm can extract accurate and effective features and obtain high classification accuracy, thus verifying the effectiveness of the proposed algorithm. Moreover, the network structure of the proposed algorithm is relatively simple with a low computation cost, which makes it feasible of further extension to the classification application of other cancer cells.

**Keywords** Feature extraction · CART feature selection · PSO-SVM · white blood cell classification.

## 1 Introduction

According to the global pathology report [1], the incidence of blood cancer has increased year by year, with more than 500,000 new patients each year. However, the method of checking blood cancer by manual reading has a low efficiency and the classification is costly [2]. Therefore, the computer-based blood cancer classification methods are increasingly being a research hotspot.

Cellular features are the basis for the classification of cells by cytopathologists. The selected cell features must not only match the diagnostic experience of the physician but also be entered in a format understood by computers. In [3], the features of blood cells were analysed in detail, and a total of 87 features were extracted. However, different features have different effects on the accurate classification of cells. Accurate and effective feature selection can not only improve the classification speed, but also improve the classification accuracy and reliability [4].

There are two main methods of feature selection, namely the Filter method and the Wrapper method. The Filter method converges to the important features quickly, but it is not related to the subsequent classification algorithm, which does not consider the influence between the feature subset and the classification algorithm. The Wrapper method has higher classification accuracy and the features subset is of higher quality, however, this method converges slowly due to the large amount of data sets. Y. Marinakis [5] used genetic algorithms to select several features with the best classification ability among 20 features of blood cell images. Duan [6] used rough sets to select features of multi-mark data, and a series of experiments proved the effectiveness of the algorithm.

In addition, the embedding method is an emerging feature selection algorithm, which combing the feature selection algorithm and the classification algorithm to form a learning model. The feature selection is automatically completed in the classification process, which is characterized by interaction with the classification algorithm. The main embedding methods are support vector machine recursive feature elimination method [7], weighted pure bayesian method [8], boolean function method [9], free forest [10] and decision tree algorithm. CART [11] is a typical binary tree in decision trees and it is widely used in medical analysis. Chen [12] used CART algorithm to obtain a CART model on lung cancer data with excellent classification ability. Kong [13] used CART algorithm in the classification of breast cancer to prevent and reduce the occurrence of breast cancer. Gasparoviga-asite [14] used the CART algorithm to reduce the characteristic dimension of proteins and selected the features. In addition, CART is widely used in other fields. Li [15] used CART algorithm in financial risk forecasting of listed companies, verifying that the algorithm is efficient. Although the CART algorithm is recognized as a classification algorithm, the classification effect for different features cannot be guaranteed. Therefore, choosing a fast and efficient feature selection algorithm will be more conducive to cell classification.

In the study of blood cell classification, there are many features and they are multi-dimensional in the images. SVM has excellent adaptability to process data in the high-dimensional space, meanwhile it achieves good performance when the data dimension is large [16]. Chen [17] used SVM to classify Pap-Smear cells and the cells were classified into four categories based on the degree of canceration. Kumar [18] used SVM to divide the blood cells into normal and abnormal cells, whose classification accuracy was 85.19%. Plissiti [19] used SVM to pinpoint the nucleus with excellent cell classification. Although the traditional SVM model is already good, there are still some problems in solving the classification problem. For example, the penalty factor $c$ and the radial basis function parameter $\sigma$ affect the classification performance of SVM [20]. Therefore, this paper introduces PSO to optimize the above two parameters and builds a PSO-SVM model, which can achieve a better performance.

To alleviate the under-fitting caused by high-dimensional cell features and improve the efficiency and accuracy of blood cell, this paper proposes a blood cell classification method based on CART feature selection. In the section 1, the morphological features, color features and texture features of blood cells are extracted. Then feature selection method is used to select 9 more effective features from 20 features in section 2. In the section 3, the PSO-SVM is used to train the cell data set, so as to achieve the correct classification of blood cells. Comparison of the evaluation indicators of blood cells is shown in section 4. Finally section 5 is the summary.

## 2 Feature extraction of bloodcells

The aim of feature extraction is to find the effective features from cell images. Some differences can be observed visually (such as colour and size), while some requires certain transformations, such as histograms and textures. Effective feature extraction directly affects the classification accuracy of cells. In general, features are extracted from the following three aspects:

(1) Colour features: this mainly refers to the colour of the nucleus or cytoplasm. After staining, the cell's nucleus and cytoplasm become different colours, making them easier to observe. In general, the extraction of colour features is mainly based on RGB colours [21], the HSI colour space is also widely used. Therefore, based on this, six colour features are selected in this paper: mean, variance, slope, kurtosis, energy, and entropy.

(2) Morphological features: there are many types of parameter extraction of morphological features. After careful observation of blood cells by Jantzen. J [22], 20 features were proposed, including area, perimeter, etc. There are many other features that can be used to describe cells, such as the nuclear-cytoplasmic ratio [23]. In this paper, eight connected chain codes are used to extract six morphological features of the cell: area, perimeter, aspect ratio, circularity, rectangularity, and nuclear–cytoplasmic ratio.

(3) Texture features: the gray-level co-occurrence matrix (GLCM) [24] is a frequently used method when extracting texture features of cells. Chen [25] judged the quality of cells by extracting the texture feature parameters of the breast tumor cells. Ning [26] extracted five texture features of the nucleus and achieved an accuracy of 88.93%. Plissiti [27] used local binary pattern features to analyse the texture features of blood cells. Given the above, the GLCM algorithm is used in this paper to extract five texture features: energy, entropy, moment of inertia, correlation and inverse moment. In addition, compared with other texture feature methods, it is more intuitive for the Tamura texture feature to obtain the visual effect corresponding to texture feature in psychology. Accordingly, three features from the Tamura feature are selected in this paper: roughness, contrast, and direction.

## 3 Feature selection of blood cells

Feature selection algorithm selects the most representative features. Liu [28] summarised the feature selection by selecting a set of optimal feature subsets, which are required to ensure the minimum number of features and the accuracy of cell recognition. Therefore, the CART algorithm is used to reduce the dimensionality of the data. The essence of CART is a binary tree, which divides the data into two parts at the node, as shown in Fig .1.

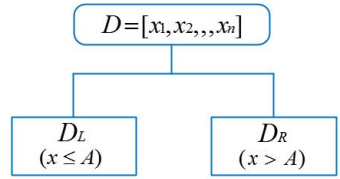

**Fig .1** Binary tree split

In each node, it is necessary to confirm a feature and its feature value to split the data. In order to measure the importance of each feature, CART uses the Gini index to observe the impracticality of the split. The Gini index is calculated as follows:

$$Gini(t) = 1 - \sum_k p_k{}^2 \tag{1}$$

where $t$ is the node of the decision tree. According to *Eq.* (1), there are k categories in the sample set, and $p_k$ represents the probability that the selected sample belongs to the k category. It can be found that the Gini index indicates the possibility of a random sample being misjudged in the subset. Assuming that the sample set corresponding to the parent node is D, which is divided into two child nodes according to feature A, and the corresponding sets are $D_L$ and $D_R$. The split Gini index is defined as follows:

$$G(D, A) = \frac{|D_L|}{D} Gini(D_L) + \frac{|D_R|}{D} Gini(D_R) \tag{2}$$

where $|\cdot|$ indicates the number of records in the sample set.

The parameter Gini_Gain is used to indicate the importance of the feature. This parameter is based on the difference between the parent node's Gini index and the child node's split Gini index when feature A is used as the split attribute. The Gini_Gain is defined as follows:

$$Gini\_Gain = G(t) - G(D, A) \tag{3}$$

Actually , Gini_Gain represents the decrease of the Gini index after each split. The larger the value, the more important the feature is. Therefore, the maximum Gini_Gain should be selected, because the essence of feature selection is to reduce the degree of uncertainty.

The flowchart of feature selection and classification for CART is shown in the Fig. 2 below.

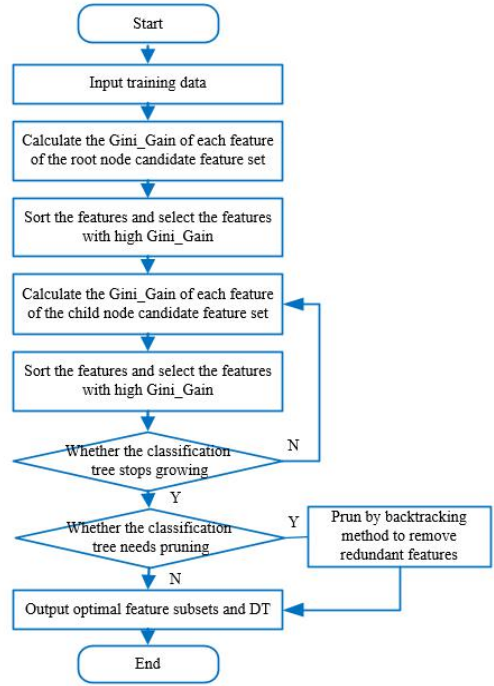

**Fig. 2** Flow chart of feature selection and classification training for CART

From Fig. 2, it can be found that the essence of the CART algorithm is to integrate feature selection and classification algorithm, which means the feature selection is automatically completed during the classification process. Although this method can extract important features, the accuracy of the classification is not enough. So this paper selects the CART algorithm for feature selection, and then uses PSO-SVM as the classifier. Therefore, important features are selected by CART algorithm, the cell classification accuracy is obtained by PSO-SVM.

Therefore, the feature selection from the 20 features is conducted, and the importance performance of 20 features in two-classification is shown in Fig. 3.

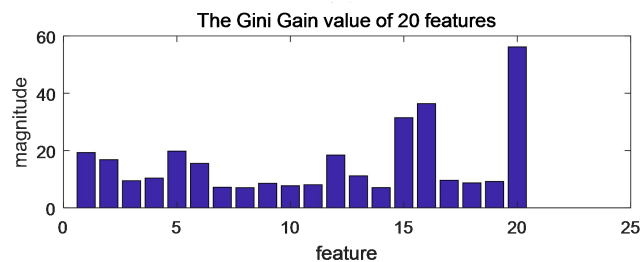

**Fig. 3** The Gini_Gain value of 20 features in two-classification

Next, the seven-classification are compared, the importance performance of 20 features in seven-classification is shown in Fig. 4.

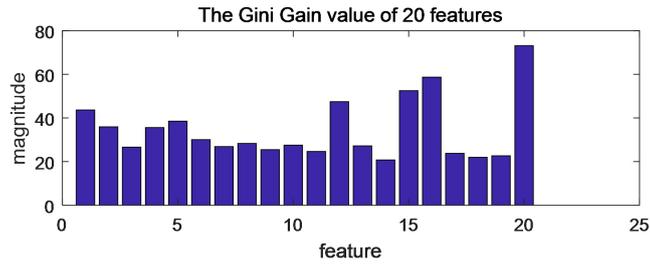

**Fig. 4** The Gini_Gain value of 20 features in seven-classification

The two metrics in Fig. 3 and Fig. 4 indicate that the feature performance of the seven-classification is almost identical to the two-classification, for which the difference between normal cells, and the difference between abnormal cells in two-classification and seven-classification are very small. The difference between normal cells and abnormal cells in two-classification is similar to the difference between the columnar epithelial cells and the mild cancerous cells in the seven-classification, thus the performance in Gini_Gain is similar.

To sort the importance of features more effectively (to observe each feature), the performance of the 20 features is sorted, and the more intuitive accuracy is as follows:

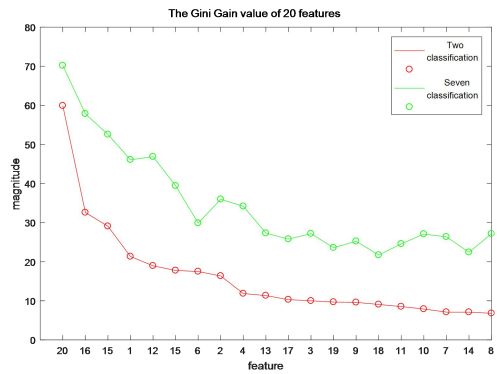

**Fig. 5** Performance of 20 features in Gini_Gain

According to Fig. 5, it can be found that the Gini_Gain of the first six features are consistent in the order of both two-classification and seven-classification. In addition, the feature 20 has the largest proportion and the effect is more obvious than others, followed by feature 16, and so on. In the seven-classification, although the performance of feature 6 is not as excellent as features 2 and 4, it occupies an important performance in two-classification; hence, its importance can be determined. By observing feature 13 and subsequent features, the reduction of these features has little effect on the Gini_Gain. Considering the performance of each feature comprehensively, the first nine features are selected and sent to the classifier for training, namely mean, variance, kurtosis, energy, entropy, roughness, perimeter, area, nuclear-cytoplasmic ratio. And they are used for subsequent training.

## 4 PSO-SVM classifier training

The SVM approach maps linearly inseparable samples to a high-dimensional linear space by using kernel functions so as to make them linearly separable, meanwhile, the computational complexity and data dimensionality can be effectively reduced [29]. The SVM achieves data classification by finding an optimal hyperplane, which can maximize the difference between classes. Take two-classification as an example:

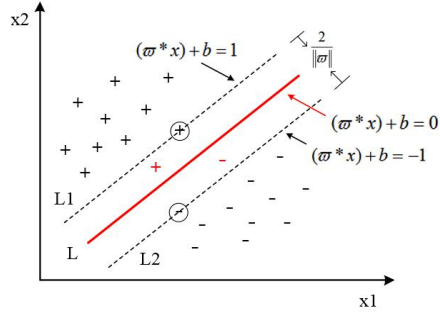

**Fig. 6** Two-classification example

Suppose there is a sample set here, the data in the sample set is $\{(x_1, y_1), (x_2, y_2), ...(x_n, y_n)\}$, $x_n \in R^d, y_n \in \{m, n\}, m \neq n$, and it is assumed that this hyperplane exists as $\varpi * x + b = 0$, then the following equation can be obtained to represent the classification:

$$(\varpi * x) + b = 1$$
$$(\varpi * x) + b = 0 \qquad (4)$$
$$(\varpi * x) + b = -1$$

Through calculating the distance between the classification lines, we discovery that the distance from the points on L1 and L2 to the classification line L are both $\dfrac{1}{\|\varpi\|}$, then the distance between L1 and L2 is $\dfrac{2}{\|\varpi\|}$. In order to maximize the difference between classes, the maximum value of $\|\varpi\|^2$ is obtained according to the formula and has the following constraint:

$$y_i(\varpi x_i + b) - 1 \geq 0, i = 1, 2, ..., n \qquad (5)$$

Therefore, under the premise of this constraint, the optimal hyperplane can be constructed initially:

$$\min(\varpi, b) = \frac{1}{2}\|\varpi\|^2 \qquad (6)$$

By this time, the Lagrange function needs to be introduced, and the following formula is obtained:

$$L = \frac{1}{2}\|\varpi\|^2 - \sum_{i=1}^{n} \alpha_i y_i(\varpi x_i + b) + \sum_{i=1}^{n} \alpha_i \qquad (7)$$

where $\alpha_i$ is the Lagrange coefficient.

In conclusion, the optimal classification function can be obtained:

$$\begin{cases} \min \dfrac{1}{2}\varpi^T \varpi + c \sum_{i=1}^{l} \xi_i \\ st. y_i(\overset{\rightarrow T}{\varpi} \phi(x_i) + b) \geq 1 - \xi_i \end{cases}$$

$$f(x) = \text{sgn}(\sum_{i=1}^{l} \alpha_i K(x_i, x) + b) \tag{8}$$

Where, c> 0 is a custom punishment factor, which reflects the degree of punishment for wrongly divided samples. $\xi_i \succ 0, i = 1, \ldots, l$ represents the number of samples. $K(x, y)$ is a kernel function, and the radial basis function is selected as the kernel function. The function formula is as follows:

$$K(x, y) = \exp(\frac{-\|x - y\|^2}{2\sigma^2}) \tag{9}$$

Although the traditional SVM model achieves good results, there are still some difficulties in selecting the hyper parameters $c$ and $\sigma$. This paper introduces the PSO algorithm to optimise the hyper parameters of SVM, making the classification results better through global optimization.

The mathematical description of PSO [30] is as follows: assume that a search space is M-dimensional, and the number of particles is n. The position of the i-th particle is expressed as

$$x_i = (x_{i1}, x_{i2}, \cdots x_{iM}), \text{ i=1,2,}\cdots\text{n,} \tag{10}$$

The fitness value can be calculated by bringing $x_i$ into the objective equation. The quality of the particles can be determined by the size of the fitness value.

The optimal position for which this paper searches is expressed as

$$p_i = (p_{i1}, p_{i2}, \cdots p_{iM}), \tag{11}$$

And the optimal position found by all particles is expressed as

$$p_g = (p_{g1}, p_{g2}, \cdots p_{gM}). \tag{12}$$

To achieve a better classification result, the dataset of this paper is divided into three parts, namely training, testing and verification sets. The specific PSO-SVM algorithm is described as follows:

(1) The value of the penalty factor $c$ and the radial basis function parameter $\sigma$ are initially determined empirically in the initial state.

(2) PSO initialization: In general, the search space is 2 dimensions, and the number of particles is between 20 and 50. Initialise the particle parameters randomly and form a particle swarm. Randomly generate the particle's starting velocity, and determine the parameters $p_i$ and $p_g$.

(3) SVM training: Calculate the fitness value of each particle with the fitness function. Update the optimal positions of the *i-th* particle and all particles at any time. If the current fitness value of *i-th* particle performs better than that of previous $p_i$, the value of $p_i$ is substituted by the current position. Being similarly, the value of $p_g$ is updated by comparing the current and previous fitness. Then $c$ and $\sigma$ are iteratively optimised to obtain the optimal parameters values.

(4) Observe whether the result achieves the required accuracy, if yes, continue the algorithm; otherwise, go to step (3).

(5) Add the verification set to observe the result, and calculate the error and fitness function. If the algorithm reaches the termination condition, continue the algorithm; otherwise, update the position of the particle, and return to (3). The termination condition of the algorithm is to achieve classification accuracy or iteration numbers.

(6) Add the testing set for test and output the classification result.

A simple PSO-SVM training flowchart is shown in Fig. 7.

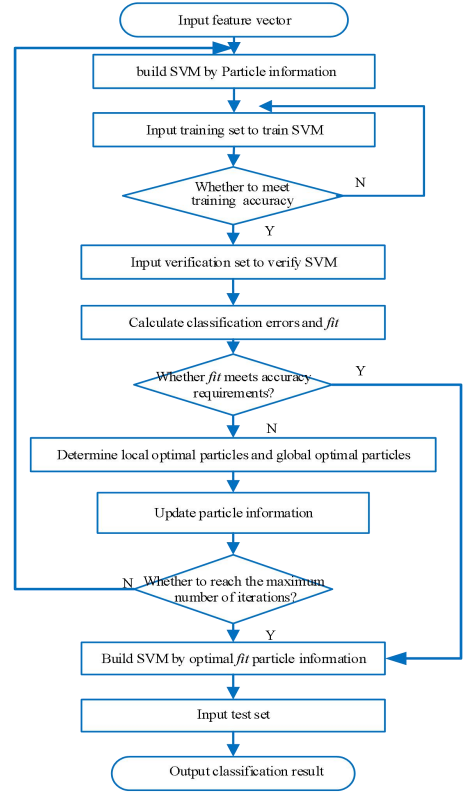

**Fig. 7** PSO-SVM model

Through the PSO-SVM algorithm, the optimal values of the parameters $c$ and $\sigma$, which minimise the SVM error, are obtained. In this paper, the training effect is best when $c = 10$ and $\sigma = 3.16$.

## 5 Comparison of the evaluation indicators of bloodcells

The results of both 20 and 9 features are performed respectively, and the comparison of two-classification and seven-classification is shown in Table 2.

**Table 2** Comparison of 20 features and 9 features in two-classification and seven-classification

| category | Two-classification | | Seven-classification | |
|---|---|---|---|---|
| | 20 features | 9 features | 20 features | 9 features |
| accuracy | 99.89% | 99.78% | 99.56% | 99.35% |
| Correctly identified number | 916 | 915 | 913 | 911 |
| Running time | 7.15s | 4.46s | 22.70s | 13.55s |

Observing Table 2, it can be found that in the process of reducing the features' number, the accuracy of cell classification is almost the same as before, but the running time of two-classification is reduced by approximately 3s, and the classification time of seven-classification is reduced by approximately 9s. The effectiveness of this feature selection method is clear.

In order to better observe the correctly identified number and the accuracy of each category in 9 features, this paper makes Tables 3 and 4 for comparison.

**Table 3** Accuracy of Two-classification

| category | numbers | Correctly identified number | accuracy |
|---|---|---|---|
| Normal cells | 242 | 241 | 99.59% |
| Abnormal cells | 675 | 674 | 99.85% |

By observing the accuracy of two-classification, one normal cell is misclassified as the abnormal cell, and one abnormal cell is misclassified as the normal cell. In the seven-classification, it can be found that four normal columnar cells are classified as mild cancerous cells, and two mildly diseased cells are classified as normal columnar cells. However, identifying normal cells as cancer cells does not cause real harm to the patient, so the classification accuracy is 99.78% for which only two cells are misidentified.

To verify the effectiveness of the proposed algorithm, principal component regression analysis (PCR) [31], principal component analysis (PCA) [32] (PCA-PSO-SVM), kernel principal component analysis (KPCA) [33] (KPCA-PSO-SVM), minimum redundancy maximum correlation (MRMR) [34] (MRMR-PSO-SVM) and ReliefF [35] (ReliefF-PSO-SVM) are used for comparison. Moreover, root mean square error (RMSE), average absolute error (AAE), maximum absolute error (MAE), Accuracy (ACC), sensitivity (SEN) and specificity (SPE) are used as the evaluation indicators, of which RMSE and ACC are the main indicators. RMSE, AAE, and MAE are defined as follows:

$$RMSE = \sqrt{\frac{1}{N}\sum\nolimits_{1}^{N}(\hat{y} - y_i)^2} \tag{13}$$

$$AAE = \frac{1}{N}\sum\nolimits_{i=1}^{N}\left|y_i - \hat{y_i}\right| \tag{14}$$

$$MAE = \max(\left|y_i - \hat{y_i}\right|) \tag{15}$$

where: $y_i$ is the true value of the *i-th* sample, $\hat{y_i}$ is the predictive value of the *i-th* sample, and $N$ is the number of the test set. The evaluation indexes are counted on the basis of reducing the number of features to nine. The performance of each specific algorithm is shown in Tables 5 and 6.

**Table 5** Comparison of test errors of different methods in two-classification

| Methods | AAE | RMSE | SPE | ACC | SEN | MAE |
|---|---|---|---|---|---|---|
| PCR | 0.1221 | 0.3495 | 83.06% | 87.79% | 89.48% | 1 |
| PCA-PSO-SVM | 0.0731 | 0.2703 | 85.95% | 92.69% | 95.11% | 1 |
| KPCA-PSO-SVM | 0.0229 | 0.1513 | 95.04% | 97.71% | 98.67% | 1 |
| MRMR-PSO-SVM | 0.0436 | 0.2089 | 92.56% | 95.64% | 96.74% | 1 |
| ReliefF-PSO-SVM | 0.09378 | 0.3062 | 88.43% | 90.62% | 91.41% | 1 |
| CART-PSO-SVM | 0.00218 | 0.0467 | 99.59% | 99.78% | 99.85% | 1 |

**Table 6** Comparison of test errors of different methods in seven-classification

| Methods | AAE | RMSE | SPE | ACC | SEN | MAE |
|---|---|---|---|---|---|---|
| PCR | 0.5247 | 0.7326 | 69.01% | 82.01% | 86.67% | 4 |
| PCA-PSO-SVM | 0.2352 | 0.6247 | 83.06% | 88.88% | 90.96% | 4 |
| KPCA-PSO-SVM | 0.0556 | 0.3252 | 92.98% | 96.07% | 97.19% | 3 |
| MRMR-PSO-SVM | 0.1123 | 0.4611 | 86.78% | 92.69% | 94.81% | 4 |
| ReliefF-PSO-SVM | 0.3542 | 0.6973 | 88.89% | 85.71% | 76.86% | 5 |
| CART-PSO-SVM | 0.00654 | 0.1144 | 99.70% | 99.35% | 98.35% | 1 |

In the both two-classification and seven-classification, the performance index of CART-PSO-SVM method is the best, with the highest accuracy. To make the results more intuitive, line charts are presented in Fig. 8.

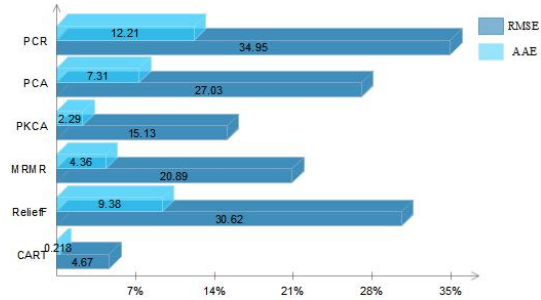

**Fig. 8**(top) Comparison of RMSE and AAE for different test methods in two-classification

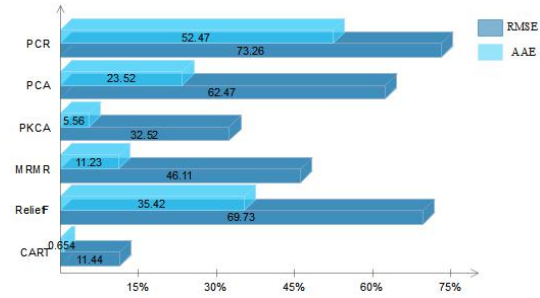

**Fig. 8**(bottom) Comparison of RMSE and AAE for different test methods in seven-classification

By observing the values of each histogram, the RMSE value of the CART-PSO-SVM achieves 3 times lower than the best performing KPCA algorithm in the other five algorithms, and the AAE value is also approximately 10 times lower than it in the two-classification. The CART-PSO-SVM can obtain lower RMSE and AAE than other five algorithm, thus its stability is well demonstrated.

The histogram of the accuracy of each algorithm in both two-classification and seven-classification is presented in Fig. 9.

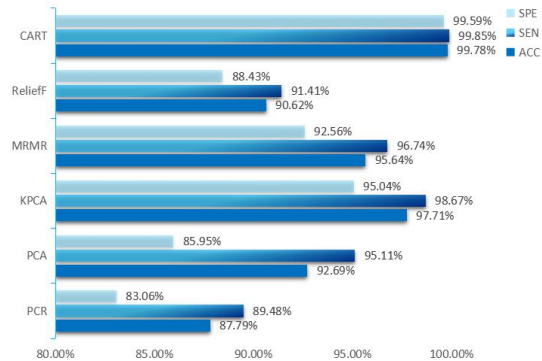

**Fig. 9**(top) Comparison of the accuracy in the two-classification

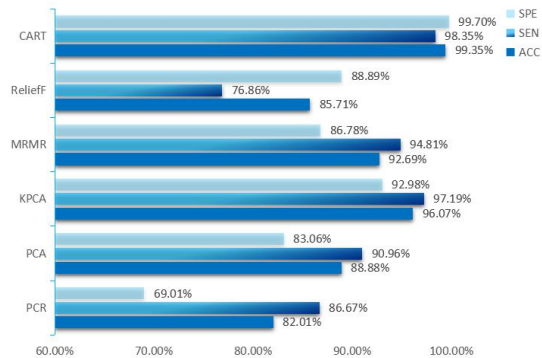

**Fig. 9**(bottom) Comparison of the accuracy in the seven-classification

From Fig. 9, it is clear that the accuracy of CART-PSO-SVM is relatively stable, with an excellent performance in both two-classification and seven-classification. The accuracy of CART-PSO-SVM is approximately 2% higher than KPCA algorithm and 12% higher than PCR algorithm in the two-classification. In the seven-classification, the classification accuracy is more obvious, which is 3% higher than the KPCA algorithm and 17% higher than the PCR. In addition, the sensitivity and specificity of the CART-PSO-SVM are also the highest.

Finally, the MAE value is specifically analysed, as shown in Fig. 10.

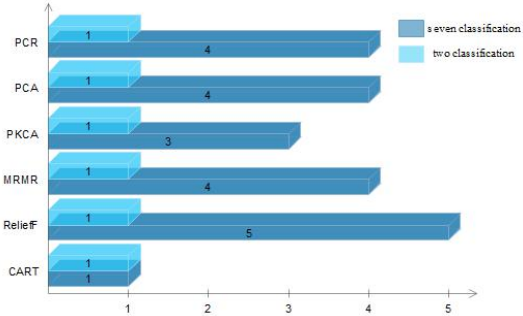

**Fig. 10** Comparison of MAE error for different test methods

Fig. 10 presents a comparison of MAE errors for different test methods. In the two-classification, the value of MAE can only be 1 as long as the classification error occurs. In the seven-classification, the larger the MAE value is, the greater the algorithm's instability is. The MAE value obtained in the ReliefF algorithm is 5, since the superficial squamous cells are misclassified as the severe dysplasia cells, which is extremely unstable. The MAE value of CART-PSO-SVM is 1, meaning only the recognition between the columnar cells and the mild dysplasia cells produces a few incorrect cells. However, the higher classification accuracy in other cancel cells is still guaranteed; thus, the superiority of the CART-PSO-SVM algorithm proposed in this paper is verified.

To verify the effectiveness of the CART-PSO-SVM algorithm more fully, the number of features and classification accuracy of several other algorithms in two-classification are listed. The specific comparison result is shown in Table 7 below.

**Table 7** Comparison of the CART-PSO-SVM algorithm and other algorithms in two-classification

|              | Accuracy | Feature numbers |
| --- | --- | --- |
| Paper [36]   | 93.41%   | 20              |
| Paper [37]   | 96.91%   | 24              |
| Paper [38]   | 95.2%    | 11              |
| Paper [39]   | 98.98%   | 11              |
| Paper [40]   | 93.72%   | 9               |
| Paper [41]   | 95.36%   | 9               |
| CART-PSO-SVM | *99.78%* | *9*             |

Among the paper [35-40], the highest classification accuracy of these algorithms is 98.98% with 11 features. The accuracy of the CART-PSO-SVM is approximately 1% higher than paper [39], achieving the best classification performance over all the other methods. Meanwhile, the number of features is reduced by 2, which effectively saves time and computation cost. It is also evident from paper [36-37] that the multi-feature classification effect is not necessarily high. Although the same nine features are selected in paper [40] and [41], the classification accuracy is 6% and 4% lower than the CART-PSO-SVM algorithm, respectively. Therefore, the number of features and the effectiveness of the selected features have a great impact on the accuracy of cell classification. Compared with the six methods, the proposed CART-PSO-SVM can achieve the highest accuracy with fewest feature numbers.

## 6 Summary

In view of the problem of feature redundancy and low accuracy in the process of blood cell classification, this paper firstly introduces CART feature selection method, which successfully reduces the extracted 20 features to 9 features, PSO is then adopted to optimise the hyper parameters of SVM algorithm, finally an efficient cell classification model with an accuracy of more than 99% is established. To verify the effectiveness of the proposed algorithm, this paper introduces RMSE, AAE, MAE and 6 other classification methods for comparison. It can be found that the accuracy of the proposed CART-PSO-SVM algorithm achieves a high level when the number of features is fewer. The result of this paper has effectively improved the classification accuracy of both two-classification and seven-classification while the computation cost is reduced and the operating efficiency is improved. It is of great application potential for reducing the misdiagnosis rate of blood cancer. At the same time, a novel framework is established for the feature selection and classification of other cancer cells, which can improve the classification speed, but also improve the classification accuracy and reliability.